# Automatic Image Annotation (AIA) of AlmondNet-20 Method for Almond Detection by Improved CNN-based Model

*Mohsen Asghari Ilani[1, *], Saba Moftakhar Tehran[2], Ashkan Kavei[3], Arian Radmehr[4]*

[1,*]*School of Mechanical Engineering, College of Engineering, University of Tehran, Tehran, Iran*
[2]*School of Electrical and Computer Engineering, University of Kashan, Kashan, Iran*
[3]*Mechanical Engineering, Islamic Azad University Science and Research Branch, Tehran, Iran*
[4]*Department of Computer Engineering, Islamic Azad University, South Tehran Branch, Tehran, Iran*

## Abstract

In response to the burgeoning global demand for premium agricultural products, particularly within the competitive nut market, this paper introduces an innovative methodology aimed at enhancing the grading process for almonds and their shells. Leveraging state-of-the-art Deep Convolutional Neural Networks (CNNs), specifically the AlmondNet-20 architecture, our study achieves exceptional accuracy exceeding 99%, facilitated by the utilization of a 20-layer CNN model. To bolster robustness in differentiating between almonds and shells, data augmentation techniques are employed, ensuring the reliability and accuracy of our classification system. Our model, meticulously trained over 1000 epochs, demonstrates remarkable performance, boasting an accuracy rate of 99% alongside a minimal loss function of 0.0567. Rigorous evaluation through test datasets further validates the efficacy of our approach, revealing impeccable precision, recall, and F1-score metrics for almond detection. Beyond its technical prowess, this advanced classification system offers tangible benefits to both industry experts and non-specialists alike, ensuring globally reliable almond classification. The application of deep learning algorithms, as showcased in our study, not only enhances grading accuracy but also presents opportunities for product patents, thereby contributing to the economic value of our nation. Through the adoption of cutting-edge technologies such as the AlmondNet-20 model, we pave the way for future advancements in agricultural product classification, ultimately enriching global trade and economic prosperity.

## 1. Introduction

Agriculture plays a critical role in a country's long-term economic health. It goes beyond simply providing food and materials. It's a major source of jobs, often the primary income source, for a significant portion of the population. The agricultural landscape is constantly changing due to various factors. Rising incomes, globalization, and increased focus on healthy eating all influence how food is produced. In the coming years, the demand for various food items like fruits, vegetables, dairy, seafood, and meat is expected to rise significantly. Developing nations like Iran face a particular challenge in agriculture: a lack of automation and mechanized processes. Despite this, a substantial portion of Iran's population (around 58%) relies on agriculture for their livelihood. The country has significant potential for growth in the food processing sector, which could position them well for increased participation in global food trade. Iran boasts a thriving grocery and food market, with retail sales accounting for a remarkable 70% of total revenue. The food processing industry is also a major contributor, holding a significant 32% share of the country's entire food market [1]. The food industry operates in a cutthroat environment where quality is the ultimate differentiator. Consumers are no longer satisfied with the bare minimum. Their heightened awareness and evolving preferences demand a relentless pursuit of excellence [2]. Despite advancements in technology, a significant portion of food quality evaluation remains stubbornly manual. This reliance on human inspectors, while possessing valuable experience, introduces a layer of subjectivity and inconsistency. Physiological factors like fatigue or even hunger pangs can influence their judgment, leading to fluctuations in the evaluation process. This traditional method also proves to be inefficient. Manual inspections are time-consuming and labor-intensive, ultimately driving up production costs. To stay ahead of the curve and meet the ever-increasing demands of today's consumers, the food industry requires a more robust and objective quality evaluation system. Here's where automation steps in, offering a compelling solution [3, 4].

Our increasingly globalized world relies heavily on the complex machinery of the food industry. This vast system encompasses everything from the planting of seeds (agriculture) to the products on our supermarket shelves (food processing, marketing, and sales). It's a tirelessly working engine ensuring a steady flow of food across the globe. However, a crucial element within this industry, food quality evaluation, remains surprisingly reliant on a traditional method: manual assessment by trained individuals. While this approach has served the industry for a long time, it comes with inherent limitations. Manual evaluations are expensive, requiring a significant workforce. Additionally, they are inherently subjective, as human inspectors rely on their own perceptions, which can be influenced by factors like fatigue or even personal preferences. This subjectivity can lead to inconsistencies in the evaluation process, potentially allowing products that don't meet quality standards to slip through the cracks [1]. As a result, the food industry faces a growing pressure to elevate its standards of food quality evaluation. The need for objectivity, consistency, and efficiency in this critical process is becoming increasingly important. The global food industry is a complex and multifaceted giant, weaving together agriculture, food processing, marketing, and sales to bring food from farm to table. Its importance in our interconnected world is undeniable. Yet, a surprising truth lies at the heart of this industry: food quality evaluation often remains a stubbornly manual process, relying on trained individuals. While these inspectors bring valuable experience, this traditional method suffers from several drawbacks. It's a costly endeavor, requiring significant manpower. More importantly, it's inherently subjective. Human inspectors, susceptible to fatigue, personal preferences, or even hunger pangs, can introduce inconsistencies into the evaluation process. This subjectivity creates a risk‐products that don't meet quality standards could slip through the cracks. As a result, the industry faces a growing demand for a more robust system‐one that prioritizes objectivity, consistency, and efficiency in food quality evaluation [1]. Almonds (Amygdalus Communis L.) are a fascinating nut, playing a dual role in the world of food and agriculture. Not only are they a valuable source of health benefits, but they also contribute significantly to thriving export industries in many regions. These perennial plants, cultivated in cold and temperate areas, produce energy-packed kernels. Aydin (2003) highlights the impressive nutritional profile of almonds, boasting 6 kcal g-1 of energy, 15.64% protein, and a remarkable oil content ranging from 35.27% to 40%. Notably, almond kernel oil stands out for its high concentration of oleic acid, clocking in at around 40 % [5].

Iran is a major player in the global almond production scene. However, their harvesting and handling methods currently rely heavily on manual labor [6]. Practices like threshing are often done by hand or with the aid of simple, homemade equipment. To optimize these various processes‐threshing, conveying, sorting, and storing ‐ a deeper understanding of the physical and mechanical properties of both the almond nuts and their kernels is crucial. This knowledge can lead to significant improvements in efficiency and overall crop management. To improve efficiency throughout the almond production chain, from harvesting to storage, researchers are actively investigating the nuts' physical and mechanical properties. One study explored the relationship between size and rupture strength across ten different almond varieties [7]. Another delved into how moisture content affects the physical properties of both almond nuts and kernels. Additionally, research has shed light on how factors like irrigation regimes, fertilization types, and even the cultivation year itself can influence the physical properties of almonds [6]. By understanding these various aspects, researchers can develop strategies to optimize processes like threshing, conveying, sorting, and storing, ultimately leading to a more efficient and productive almond industry. Currently, the process of grading almonds relies heavily on manual methods. One traditional technique involves calculating "adjusted kernel weight." This calculation factors in the weight of edible kernels, inedible ones, foreign materials, and excess moisture, using predetermined percentages [8, 9]. While this method exists, it suffers from several drawbacks. It's time-consuming, labor-intensive, and ultimately inefficient. Additionally, it struggles with consistency, especially when evaluating visual aspects of the almonds like broken pieces, halves, chips, scratches, splits, and the condition of their shells. In an attempt to improve efficiency, computer-based databases were introduced to replace manual calculations and grading. However, even these digital systems weren't without flaws. Rounding errors, particularly at the hundredth decimal place, persisted [8]. This lack of precision could potentially lead to inconsistencies in the final grading results.

While manual methods have traditionally dominated almond grading, recent advancements in image processing and analysis offer a promising alternative. This technology utilizes digital imaging, computer scanning, and specialized software to analyze almond kernels [10–13]. The software calculates factors like kernel area and pixel values, allowing for a more objective and automated classification process. Machine learning techniques like SHAPE, SeedCount, GrainScan, Smartgrain, and ImageJ have shown potential in effectively classifying different types of kernels in critical domains, such as agriculture and healthcare [14-15]. However, there are limitations to consider. Some of these techniques are still under development, and others may be too cumbersome or expensive for large-scale applications. The key lies in finding an image processing solution that strikes a balance between accuracy, efficiency, and cost-effectiveness for the high-volume world of almond production. The world of almond grading is undergoing a fascinating transformation. While manual methods have served the industry well for many years, researchers are increasingly exploring the potential of machine learning to automate and improve the process. Several studies have shed light on this exciting development. Mirzabe et al. [33] investigated the physical properties of almonds, employing statistical models and machine learning techniques to analyze factors like bulk density and friction coefficient. Teimouri et al. [29] took a different approach, utilizing artificial neural networks (ANNs) to successfully categorize almonds into various classes, including normal, broken, wrinkled, and double kernels. Their research yielded impressive results, demonstrating the high sensitivity and accuracy of this approach in detecting almond quality variations. Sharifi et al. [34] explored the integration of artificial intelligence with digital twin models to optimize stormwater management in smart cities, highlighting the enhanced efficiency and sustainability of urban drainage systems. Lonbar et al. [35] investigated risk factors using hybrid fuzzy Z-number approaches for strategic budget allocation in media and construction campaigns.

Another study (Vidyarthi et al. [8]) combined image processing with machine learning models like random forest and support vector machines (SVM) to predict the size and mass of almond kernels. This research highlights the versatility of machine learning, showcasing its potential application beyond simple classification tasks. Additionally, Reshadsedghi et al. [16] explored the use of ANNs to classify almond varieties based on shell characteristics. Their approach involved extracting acoustic features from the shells and feeding them into the ANNs, achieving high classification accuracy for different shell types. Eski et al. [17] further broadened the scope of machine learning applications in almond production by designing a system to predict various physical properties based on just three key dimensions – length, width, and thickness. This research demonstrates the potential of machine learning to extract valuable insights from limited data sets. While these studies showcase the immense potential of machine learning in almond grading, it's important to acknowledge that some techniques are still under development. Additionally, cost-effectiveness remains a crucial consideration, particularly for large-scale production. Nevertheless, the research is clear: machine learning holds the promise to revolutionize almond grading, ushering in an era of greater efficiency, accuracy, and consistency. Almond recognition and quality assessment model presented in this study demonstrate promising real-time outcomes and can seamlessly integrate with camera-based systems for on-the-fly fruit quality analysis. Notably, the AlmondNet-CNN based architecture is characterized by a reduced number of parameters, enhancing its efficiency in training on a large volume of images within a shorter timeframe. Consequently, the model's processing time for real-world images is minimized, making it highly suitable for precision agriculture applications. As the world of almond grading seeks to move beyond traditional manual methods, the integration of digital image processing and pattern recognition algorithms has emerged as a game-changer [18]. Ahmadi et al. [20] conducted an assessment of SAM and U-Net deep learning architectures for the identification of concrete cracks. SAM efficiently pinpoints longitudinal cracks using image segmentation techniques, while U-Net precisely identifies spalling cracks by analyzing the characteristics of pixels. Employing both models together significantly enhances the detection of cracks, which is crucial for the safety and longevity of concrete structures. Here, computer vision steps in as a powerful tool, offering speed, cost-effectiveness, consistency, and most importantly, precision in inspection tasks. The food industry, in particular, has witnessed a surge in the application of computer vision for quality assessment. Recognized for its vast potential, computer vision has placed the food industry among the top 10 sectors actively employing this technology [8]. Its success lies in its ability to perform objective and non-destructive evaluations across a wide variety of food products.  This has been a key driver for substantial research and development efforts within the food industry, as the advantages of computer vision - objectivity, speed, and the contactless nature of inspection - have become increasingly clear [19]. Ahmadi et al. [30] created a supervised machine learning framework tailored for digital twin uses, specifically for segmenting terrain in coastal areas. Utilizing USGS datasets and advanced deep learning methods, they categorized the coastal landscape of Florida into distinct classes like water, grassland, and forest. This segmentation significantly improves digital twins for more effective environmental surveillance and urban development planning. Govindan et al. [31] developed a stochastic optimization model for creating a resilient reverse logistics network to manage infectious healthcare waste during crises like COVID-19. The model incorporates strategies like new collection centers and third-party logistics to enhance resilience and efficiency in waste management. Some papers, such as [36-41], have also been reviewed and investigated due to the practical solutions and insights they provided. In the context of computer vision for food quality assessment, a well-designed system goes beyond just the software and algorithms. A critical partnership exists between the computer vision system itself and an illumination system. Imagine a personal computer (PC) equipped with specialized software ready to analyze food products.  However, for this analysis to be effective, it needs clear and accurate information. This is where the illumination system comes in. Just like good lighting is essential for taking a clear photograph, the illumination system plays a vital role in capturing high-quality images of food products. The quality of the captured image has a significant impact on the entire process. A well-lit image allows for faster and less complex image processing steps later on. This translates to a more efficient system overall, and even plays a role in reducing the overall cost. In simpler terms, good lighting upfront means a smoother and more cost-effective computer vision system for food quality evaluation.

This research takes a groundbreaking approach to almond quality assessment by introducing a novel deep learning model called Almond-CNN. This model is specifically designed to identify and evaluate the quality of individual almonds within a mixed batch containing both almonds and shells. The researchers created a robust dataset for training and testing the model. This dataset is comprised of real-world scenario images meticulously categorized into two distinct quality classes. The core of the system lies in a convolutional neural network (CNN) architecture. To ensure optimal performance, the model is trained on this comprehensive dataset of 736 images, encompassing a variety of almond and shell combinations, over multiple training cycles (epochs). Finally, the trained deep learning model undergoes rigorous testing to validate its accuracy and effectiveness [32]. The challenges faced by the researchers in developing the Almond-CNN model were not insignificant. The dataset they compiled to train the model exhibited significant variations, both between different classes (almonds vs. shells) and even within the same class (variations among almond shapes and appearances). Additionally, the real-world scenarios these images captured added another layer of complexity. To address these challenges, the researchers divided their comprehensive dataset of 736 images into three distinct subsets: training, validation, and testing. This meticulous approach allowed them to train the model effectively while also ensuring its generalizability to unseen data. The results were impressive. The deep learning-based Almond-CNN model surpassed all existing state-of-the-art models, achieving a remarkable 100% accuracy on a test set featuring entirely new images. This speaks volumes about the model's effectiveness in real-world applications. Beyond accuracy, the researchers also focused on creating a model that could seamlessly integrate into existing workflows. The fruit recognition and quality assessment system they developed demonstrates promising real-time capabilities. In simpler terms, the model can analyze fruits as they appear, without any delays. Furthermore, the architecture of Almond-CNN is designed for efficiency. By using a reduced number of parameters, the model can be trained on large datasets in shorter timeframes. This translates to faster processing times for real-world images, making it ideal for applications in precision agriculture, where speed and efficiency are crucial.

## 2. Materials and Methods

This study investigated the potential of Convolutional Neural Networks (CNNs), a powerful deep learning technique, for recognizing and classifying almond kernels. Deep CNN architectures have become the go-to tools for effective image recognition, detection, and classification tasks due to their superior performance [21, 22]. A typical CNN architecture comprises multiple layers, including convolutional layers for feature extraction, non-linear activation layers for introducing non-linearity, and pooling layers for dimensionality reduction [23]. Deep CNNs offer several advantages over traditional shallow neural networks, including sparse interactions (limited connections between neurons), parameter sharing (reducing training complexity), and equivariance (consistent response to similar input features). These advantages make DCNNs superior classifiers compared to traditional approaches like logistic regression, linear regression, and Support Vector Machines (SVMs).

Beyond basic CNN architecture, this research incorporated advanced image processing and machine learning techniques to achieve high classification accuracy. These techniques included image augmentation (artificially creating more training data by manipulating existing images), transfer learning (leveraging pre-trained models on similar tasks to accelerate training and improve performance), and dropout (randomly dropping neurons during training to prevent overfitting). By utilizing these techniques, the DCNN model effectively mitigated data noise and interference, leading to a more robust recognition and classification system for almond kernels.  Transfer learning played a crucial role in this study. This technique involves repurposing a pre-trained model on a related task to accelerate training and enhance the performance of the new model. In this case, well-established deep learning models like AlexNet and GoogLeNet, already trained on large image datasets, provided a valuable starting point. Transfer learning significantly reduced the time and effort required to train the new DCNN model for almond kernel classification [24]. The research also focused on the effectiveness and adaptability of CNNs in preserving spatial relationships between data points within 3D image tiles. Unlike traditional neural networks where learnable parameters increase significantly with the number of input features, CNNs scale their learnable parameters with filter size and the number of filters used. This allows for increasing the number of layers without a substantial increase in learnable parameters, promoting model efficiency [22]. After experimenting with various configurations, including different layer counts, kernel sizes, and pooling options, the researchers settled on a specific CNN architecture (detailed in *Figure 4*) for subsequent training and testing. This architecture, similar to a previous strategy used for almond detection, achieved the highest average classification accuracy. The chosen CNN architecture consisted of three sets of sequential layers: 3D Convolution (CONV) for feature extraction, Batch Normalization (BN) for stabilizing training, Rectified Linear Unit (ReLU) for introducing non-linearity, and Max Pooling (MP) for dimensionality reduction. These sets of layers were followed by Fully Connected (Dense) layers for combining extracted features, a Softmax layer for generating probability distributions, and finally a Classification layer for assigning class labels (e.g., almond kernel or non-kernel) to the input data. Each convolutional layer utilized various 3D convolution filters to extract relevant features within the input dataset [16]. The resulting features were then concatenated along a new dimension, creating a richer representation for classification.

## 2.1 Image annotation

The ever-increasing volume of digital images, coupled with user demand for efficient access to vast data sets, has fueled the need for accurate and fast image retrieval technology. Automatic Image Annotation (AIA) emerges as a key solution, enabling image retrieval based on textual content. AIA presents a significant challenge in computer vision, as described by Barnard et al. [25], due to the inherent complexity and diverse nature of image content. It allows for swift and effective query, retrieval, and organization of image information. AIA has applications in various areas, including online and offline data exploration, image manipulation, and mobile annotation tools [25–28]. A typical image annotation system relies on two critical elements: (1) a mechanism for semantically understanding the content of raw images, and (2) a Natural Language Processing (NLP) unit that translates the extracted semantic data into human-readable labels [29]. In this study, we utilized LabelImg, a popular tool, for multi-class labeling of images. *Figure 1* showcases the LabelImg framework used for labeling almonds and shells. Specifically, *Figure 1* (a, b, c) depict different instances of the labeling process*Figure 1* (d) illustrates the labeling of a single almond, while *Figure 1* (e) demonstrates the labeling of shells. LabelImg generates the labeled images as XML files, which capture both bounding boxes (encompassing the objects) and corresponding labels.

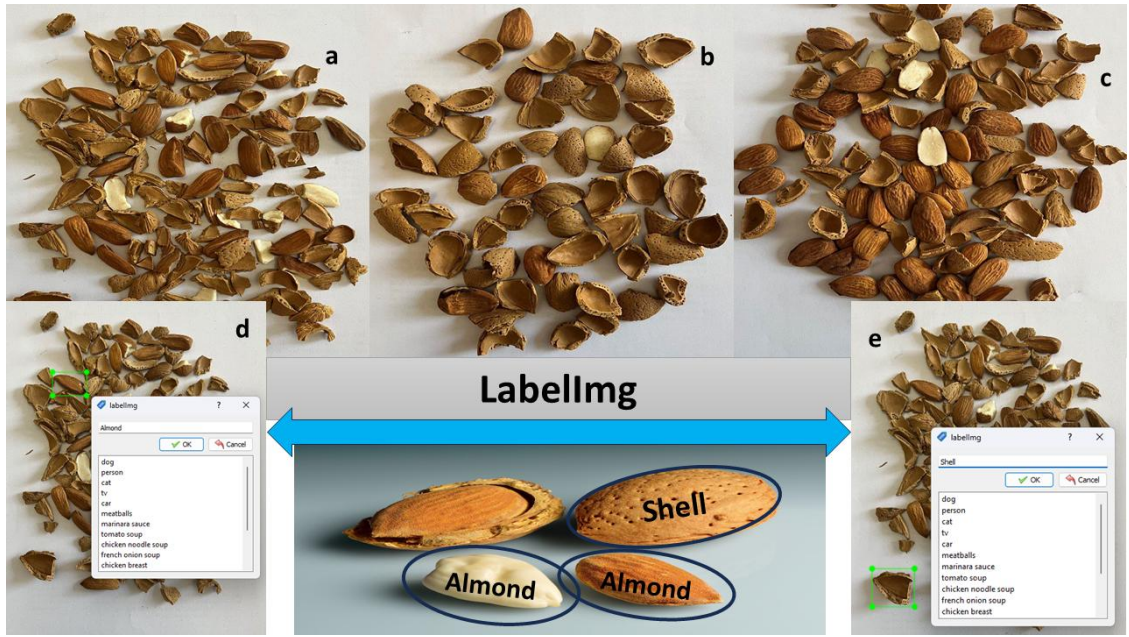

**Figure 1.** Annotation of images within the LabelImg environment. **(a-c)** Original images encompass both almonds and shells. **(d)** Almond labeling in LabelImg. **(e)** Shell labeling in LabelImg.

Labeling these processed images using LabelImg adds another layer of refinement to the model's training dataset. By precisely annotating the images with bounding boxes and labels, LabelImg contributes valuable information for training and validating the CNN model. This labeled data enhances the model's ability to recognize and differentiate between almond and shell instances during training, ultimately leading to higher accuracy and improved performance. The combination of thoughtful image processing and meticulous labeling creates a robust foundation for training the CNN model, ensuring it can effectively learn and generalize patterns, resulting in more accurate and reliable object detection outcomes.

## 2.2 Image Pre-processing

Implementing a pre-processing pipeline before feeding images into a CNN model can significantly enhance accuracy and improve results. In the case of images containing both shells and almonds, the applied image processing steps play a crucial role. Converting the RGB images (**Figure 2** (a)) to grayscale (**Figure 2** (b)) simplifies the information while retaining essential features. The subsequent application of Gaussian blur and denoising techniques (**Figure 2** (c)) helps in reducing noise and enhancing the clarity of relevant image details. Thresholding (**Figure 2** (d)) further aids in isolating distinct features by creating a binary representation, making it easier for the CNN model to focus on relevant information. Finally, employing Canny edge detection (**Figure 2** (e)) contributes to the extraction of meaningful edges, providing the model with well-defined boundaries for object recognition. This comprehensive image processing approach not only refines the input data but also optimizes the images for improved learning and classification within the CNN model.

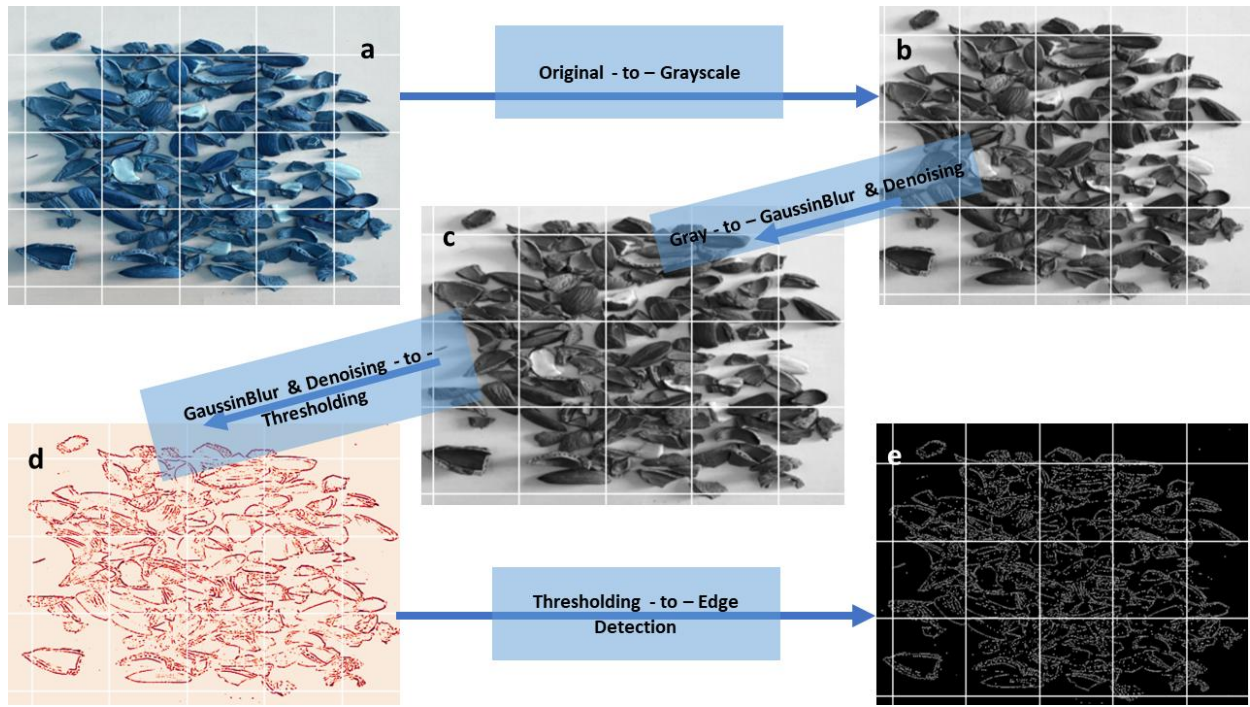

**Figure 2.** Preprocessing of labeled images with OpenCV involves several steps: (a) Original image, (b) Grayscale conversion, (c) Blurring and denoising using Gaussian and fastNlMeansDenoising, (d) Thresholding through adaptive Threshold, and (e) Edge detection using Canny.

To enhance the training of our CNN architecture, we implement an effective image balancing strategy using scikit-learn's class weight computation method (**Figure 3** (a, b)). This method assigns different weights to the classes based on their frequencies in the dataset, ensuring a balanced representation during training. The application of class weights is particularly valuable when dealing with imbalanced datasets, as it prevents the model from favoring the majority class. This process is instrumental in addressing potential biases and improving the model's ability to generalize across different classes. After computing the class weights, we prepare the images for the train-test split (**Figure 3** (c)). The balanced dataset is divided into training and testing sets, providing the CNN model with diverse examples for learning and evaluation. The goal is to prevent overfitting and enhance the model's performance on unseen data. The utilization of class weights ensures that the CNN architecture is exposed to an equitable distribution of both almond and shell instances during training, contributing to a more robust and accurate model.

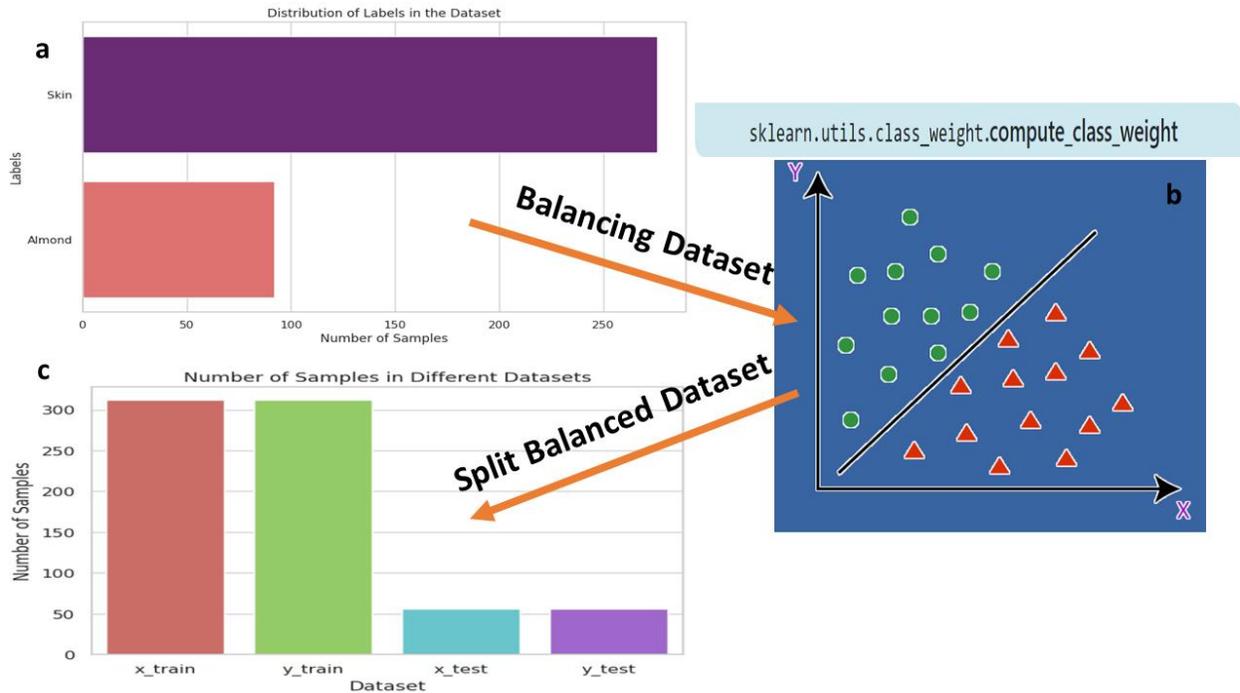

**Figure 3.** Balancing labeled images for dataset splitting involves the following steps: (a) Utilizing an unbalanced dataset with shells and almonds. (b) Achieving balance through class weights computed using the scikit-learn library. (c) Dividing the dataset into training and test datasets.

## 2.3 CNN Model Architecture

The CNN architecture is designed to effectively detect almonds and shells in images. The model, defined as a Sequential model, starts with a convolutional layer (Conv2D) with 64 filters of size 3x3, using the ReLU activation function and maintaining spatial dimensions with padding. The input shape is set to (312, 210, 320, 1), reflecting the dimensions of the images. Subsequently, max-pooling layers (MaxPooling2D) with a pool size of 2x2 are applied to down-sample the feature maps while preserving spatial information.

The architecture further incorporates a convolutional layer with 128 filters and another set of max-pooling layers. SpatialDropout2D is introduced for regularization to enhance model generalization. Additional convolutional layers with 512 and 256 filters follow, each accompanied by max-pooling. A stride of 2 is applied in the 256-filter layer for further down-sampling. The subsequent layers include another convolutional layer with 256 filters and a unique max-pooling layer with a pool size of 3x3. Two more convolutional layers with 128 filters and corresponding max-pooling layers are included, concluding with a dropout layer for regularization. Following the convolutional layers, the model transitions to a flattened layer, preparing the data for fully connected layers. A dense layer with 64 neurons and ReLU activation is added, providing a nonlinear transformation. BatchNormalization is introduced to normalize and stabilize activations, contributing to model robustness. The final layer consists of two neurons with a softmax activation function, facilitating binary classification between almonds and shells. This CNN architecture, with its combination of convolutional, max-pooling, and fully connected layers, is well-suited for effectively learning and classifying features in almond and shell images.

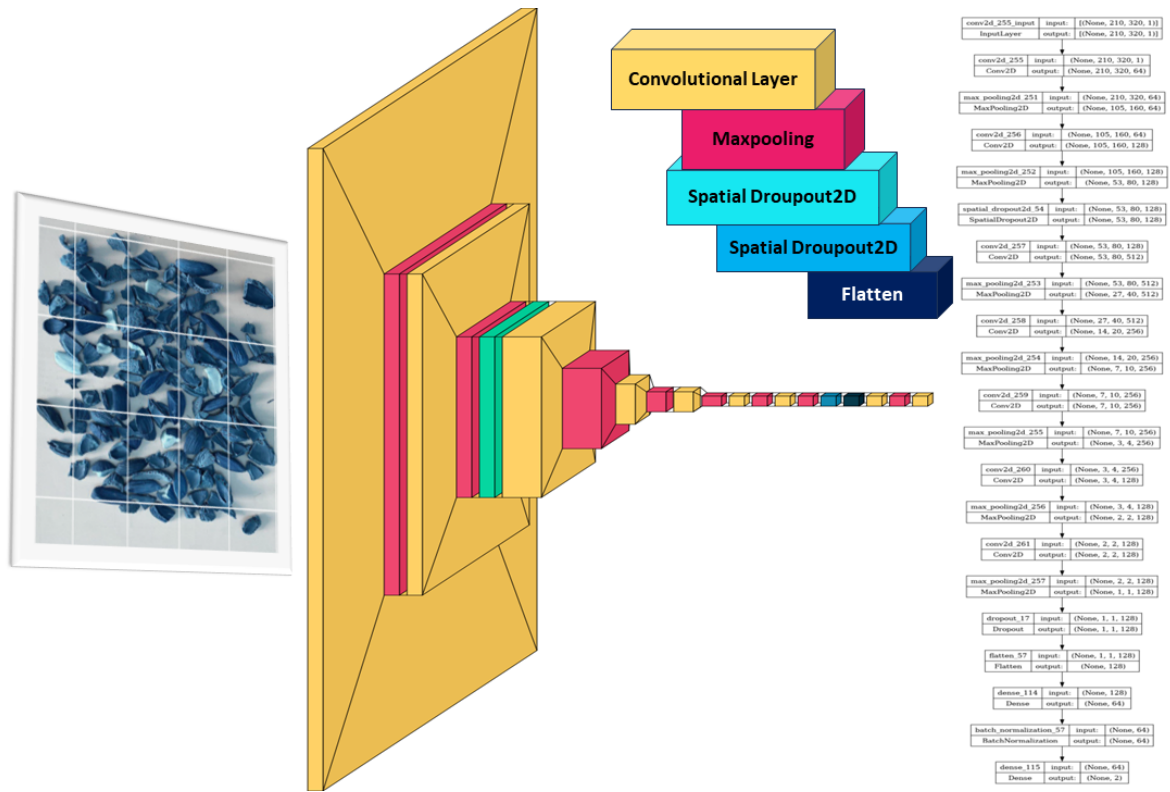

**Figure 4.** CNN-based Model Architecture.

## 3. Results and Discussion

This investigation leverages annotated objects from the LabelImg tool to enhance object identification accuracy, supplementing the model's training. Figure 5 provides a concise overview of training and validation outcomes. The dataset maintains a balanced 1:1 distribution of Almonds and Shells and intact images, with a 5:1 training-validation split. Training accuracy is computed from a dataset of 736 images, and the validation set constitutes 20% of the training data. Exceptional accuracy is achieved, reaching close to 100% during the 831st epoch for training and 97.95% at the 899th epoch for validation. The use of two GPUs accelerates training in Kaggle, reducing the time needed for the 1000th epoch by approximately 40 minutes. Notably, the estimated running time on a CPU alone exceeds 6 hours, underscoring the computational efficiency gained through GPU acceleration.

In our study, the Convolutional Neural Network (CNN) exhibited significant performance evolution over 100 epochs, evident in decreasing loss and increasing accuracy for both training and validation datasets. Initially, during the first epoch, the training data showed a loss of 1.7970 with an accuracy of 47.39%, while on the validation set, the loss was 1.4931 with an accuracy of 26.98%. Subsequent epochs displayed a consistent trend of diminishing loss values and ascending accuracy scores. The final epoch, epoch 1000, demonstrated remarkable convergence, with minimal loss values of 0.0567 on the training set (*Figure 5* (a)) and 0.0605 on the validation set (*Figure 5* (b)), accompanied by high accuracies of 100% and 100%, respectively. This convergence signifies the model's successful learning and generalization over the training period (*Figure 5* (a,b)).

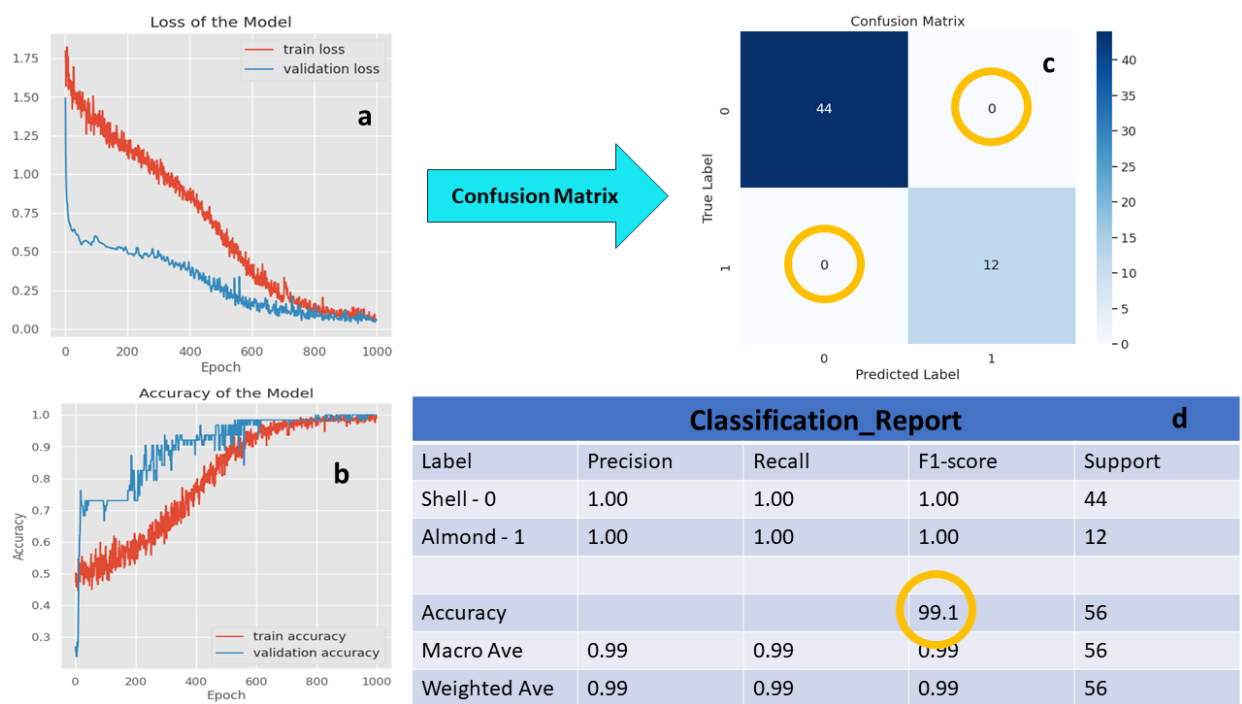

**Figure 5.** Model Results. (a) Loss of the Model. (b) Accuracy of the Model. (c) Confusion Matrix. (d) Classification Report.

The CNN model exhibited highly efficient performance in detecting various objects, as indicated by precision, recall, and F1-score metrics in *Figure 5* (d). Specifically, almond and shell detections achieved a precision, recall, and F1-score of 1.00, demonstrating robust identification. The model's overall accuracy across all classes reached 0.99, underscoring its effectiveness in Almond identification within the dataset. Micro-averaged and weighted-averaged metrics further support the model's consistent and high-level performance across different type of objects as Shells and Almond categories, with reported values of 0.991 for precision, recall, and F1-score in both micro and weighted averages. These results affirm the model's proficiency in efficiently detecting and classifying almonds and shells within the dataset comprising 106 tests.

The overall prediction accuracy is depicted through a confusion matrix in *Figure 5* (c), summarizing accurate predictions across all class labels in matrix form. The diagonal blue elements represent the number of correct predictions, with 44 for shell detection and 12 for almond detection out of the total data. White elements indicate inaccurate predictions, which are 0 for both classes.

## 4. Conclusion

In conclusion, our research presents a novel approach utilizing the LabelImg methodology to enhance the accuracy of almond detection within complex environments, leveraging the intelligence of Convolutional Neural Networks (CNNs). The intricate nature of almond and shell objects necessitates the adoption of a CNN-based approach, offering superior precision and reliability in classification tasks. By meticulously labeling images containing intact and broken almonds, we curated annotated datasets comprising bounding boxes and labels, facilitating robust training and testing phases. Furthermore, this annotated data serves as valuable feedback for refining preprocessing hyperparameters, enhancing the overall efficacy of our classification system. Our study unveils compelling findings regarding the effectiveness of deep learning methodologies in precisely categorizing almonds, with profound implications for nut packaging and the broader food and crop industries. Noteworthy key findings include the successful training of a CNN model to detect almonds amidst a high volume of almonds and shells, achieving a remarkable prediction accuracy exceeding 99.1%. Furthermore, our CNN-based model, specifically implemented on AlmondNet-20, demonstrates proficiency in distinguishing nut types, particularly almonds, boasting perfect precision, recall, and F1-score metrics. We advocate for future investigations into exploring almond types to further refine precision in classification tasks.

Moreover, the strategic balancing of datasets significantly enhances training and validation performance, manifesting in a remarkable performance improvement from 28% to 99.1%. Leveraging OpenCV for image processing, incorporating techniques such as GaussianBlur, fastNlMeansDenoising, adaptiveThreshold, and Canny, streamlines almond texture discrimination and shell identification, augmenting model accuracy and efficiency. Looking ahead, we identify LabelImg as a promising tool and strategy for handling images with intricate details, facilitating streamlined annotation processes and contributing to the overall robustness of our classification system. In summary, our research underscores the transformative potential of deep learning methodologies, particularly within the realm of agricultural product classification. By advancing almond detection accuracy and efficiency, our study holds promise for reducing time and costs associated with manual inspection and quality checks, thereby offering tangible benefits for industry stakeholders and driving advancements in agricultural technology and productivity.

**Data availability**

Due to limitations in dataset sharing, the data is available upon reasonable request. Please contact Mohsen Asghari Ilani at mohsenasghari1990@ut.ac.ir for access.

**Declarations**

**Conflict of interest**

The authors declare no conflict of interest.

# Reference

1. Kumar A, Joshi RC, Dutta MK, Jonak M, Burget R. Fruit-CNN: An Efficient Deep learning-based Fruit Classification and Quality Assessment for Precision Agriculture. International Congress on Ultra Modern Telecommunications and Control Systems and Workshops. 2021;2021-October:60–5. https://doi.org/10.1109/ICUMT54235.2021.9631643.

2. Erbaş N, Çinarer G, Kiliç K. Classification of hazelnuts according to their quality using deep learning algorithms. CZECH JOURNAL OF FOOD SCIENCES. 2022;40:240–8. https://doi.org/10.17221/21/2022-CJFS.

3. Ilani MA, Khoshnevisan M. Study of surfactant effects on intermolecular forces (IMF) in powder-mixed electrical discharge machining (EDM) of Ti-6Al-4V. International Journal of Advanced Manufacturing Technology. 2021;116:1763–82. https://doi.org/10.1007/S00170-021-07569-3/FIGURES/30.

4. Ilani MA, Khoshnevisan M. An evaluation of the surface integrity and corrosion behavior of Ti-6Al-4 V processed thermodynamically by PM-EDM criteria. International Journal of Advanced Manufacturing Technology. 2022;120:5117–29. https://doi.org/10.1007/S00170-022-09093-4/FIGURES/18.

5. Dilmac M, Altuntas E. Some physical properties of almond nut and kernel and modeling dimensional properties. International Journal of Food Engineering. 2013;8. https://doi.org/10.1515/1556-3758.2652.

6. Dursun E, Dursun I. Some Physical Properties of Caper Seed. Biosyst Eng. 2005;92:237–45. https://doi.org/10.1016/J.BIOSYSTEMSENG.2005.06.003.

7. View of Some physical properties of almond nut and kernel and modeling dimensional properties. n.d. https://cigrjournal.org/index.php/Ejounral/article/view/2311/1747. Accessed January 17, 2024.

8. Vidyarthi SK, Singh SK, Xiao HW, Tiwari R. Deep learnt grading of almond kernels. J Food Process Eng. 2021;44:e13662. https://doi.org/10.1111/JFPE.13662.

9. Federal Register :: Almonds Grown in California; Revision to the Adjusted Kernel Weight Computation. n.d. https://www.federalregister.gov/documents/2018/06/20/2018-13272/almonds-grown-in-california-revision-to-the-adjusted-kernel-weight-computation. Accessed January 17, 2024.

10. Williams K, Munkvold J, Sorrells M. Comparison of digital image analysis using elliptic Fourier descriptors and major dimensions to phenotype seed shape in hexaploid wheat (Triticum aestivum L.). Euphytica. 2012;190:99–116. https://doi.org/10.1007/S10681-012-0783-0.

11. Li D, Wang L, Wang M, Xu YY, Luo W, Liu YJ, et al. Engineering OsBAK1 gene as a molecular tool to improve rice architecture for high yield. Plant Biotechnol J. 2009;7:791–806. https://doi.org/10.1111/J.1467-7652.2009.00444.X.

12. Tanabata T, Shibaya T, Hori K, Ebana K, Yano M. SmartGrain: High-Throughput Phenotyping Software for Measuring Seed Shape through Image Analysis. Plant Physiol. 2012;160:1871–80. https://doi.org/10.1104/PP.112.205120.

13. Whan AP, Smith AB, Cavanagh CR, Ral JPF, Shaw LM, Howitt CA, et al. GrainScan: A low cost, fast method for grain size and colour measurements. Plant Methods. 2014;10:1–10. https://doi.org/10.1186/1746-4811-10-23/FIGURES/5.

14. Rahimi, M., Al Masry, Z., Templeton, J. M., Schneider, S., & Poellabauer, C. (2023). Beyond motor symptoms: Toward a comprehensive grading of Parkinson's disease severity. Proceedings of the 14th ACM International Conference on Bioinformatics, Computational Biology, and Health Informatics, Houston, TX, USA. https://doi.org/10.1145/3584371.3612988

15. Rahimi, M., Al Masry, Z., Templeton, J. M., Schneider, S., & Poellabauer, C. (2024). A comprehensive multi-functional approach for measuring Parkinson's disease severity. Applied Clinical Informatics. https://doi.org/10.1055/a-2420-0413

16. Reshadsedghi A, Mahmoudi A. Detection of Almond Varieties Using Impact Acoustics and Artificial Neural Networks. 2013.

17. Eski İ, Demir B, Gürbüz F, Abidin Kuş Z, Uğurtan Yilmaz K, Uzun M, et al. Design of Neural Network Predictor for the Physical Properties of Almond Nuts. Erwerbs-Obstbau. 2018;60:153–60. https://doi.org/10.1007/S10341-017-0349-3.

18. Chen A, Nagar Y, Shoshani G, Sharon R. Deep learning for almond fruit detection at different growth stages for orchards pest management using unmanned aerial vehicles. Wageningen Academic; 2023.

19. Cruz AC, Magana S, Greco D, De Bellis L, Luvisi A. Detection of Almond Leaf Scorch with Artificial Intelligence for the Agriculture Industry. Proceedings - 2022 5th International Conference on Artificial Intelligence for Industries, AI4I 2022. 2022;1–4. https://doi.org/10.1109/AI4I54798.2022.00007.

20. Ahmadi, M., Lonbar, A. G., Sharifi, A., Beris, A. T., Nouri, M., & Javidi, A. S. (2023). Application of segment anything model for civil infrastructure defect assessment. arXiv preprint arXiv:2304.12600.

21. Lecun Y, Bengio Y, Hinton G. Deep learning. Nature 2015 521:7553. 2015;521:436–44. https://doi.org/10.1038/nature14539.

22. Rawat W, Wang Z. Deep Convolutional Neural Networks for Image Classification: A Comprehensive Review. Neural Comput. 2017;29:2352–449. https://doi.org/10.1162/NECO_A_00990.

23. Zhang Z, Jin Y, Chen B, Brown P. California almond yield prediction at the orchard level with a machine learning approach. Front Plant Sci. 2019;10:444608. https://doi.org/10.3389/FPLS.2019.00809/BIBTEX.

24. Lecun Y, Bengio Y, Hinton G. Deep learning. Nature. 2015;521:436–44. https://doi.org/10.1038/NATURE14539.

25. Barnard K, Duygulu P, Forsyth D, Blei DM, Jordan JORDAN MI, Kandola J, et al. Matching Words and Pictures Nando de Freitas. Journal of Machine Learning Research. 2003;3:1107–35.

26. Shih CW, Chu HC, Chen YM, Wen CC. The effectiveness of image features based on fractal image coding for image annotation. Expert Syst Appl. 2012;39:12897–904. https://doi.org/10.1016/J.ESWA.2012.05.003.

27. Uricchio T, Ballan L, Seidenari L, Del Bimbo A. Automatic image annotation via label transfer in the semantic space. Pattern Recognit. 2017;71:144–57. https://doi.org/10.1016/J.PATCOG.2017.05.019.

28. Ma Y, Liu Y, Xie Q, Li L. CNN-feature based automatic image annotation method. Multimed Tools Appl. 2019;78:3767–80. https://doi.org/10.1007/S11042-018-6038-X.

29. Teimouri, N., Omid, M., Mollazade, K., & Rajabipour, A. (2016). An artificial neural network-based method to identify five classes of almond according to visual features. Journal of Food Process Engineering, 39(6), 625-635.

30. Ahmadi, M., Gholizadeh Lonbar, A., Nouri, M., Sharifzadeh Javidi, A., Tarlani Beris, A., Sharifi, A., & Salimi-Tarazouj, A. (2024). Supervised multi-regional segmentation machine learning architecture for digital twin applications in coastal regions. Journal of Coastal Conservation, 28(2), 44.

31. Govindan, K., Fard, F. S. N., Asgari, F., Sorooshian, S., & Mina, H. (2024). Designing a resilient reverse network to manage the infectious healthcare waste under uncertainty: A stochastic optimization approach. Computers & Industrial Engineering, 194, 110390.

32. Hosseini Rad R, Baniasadi S, Yousefi P, Morabbi Heravi H, Shaban Al-Ani M, Asghari Ilani M. Presented a Framework of Computational Modeling to Identify the Patient Admission Scheduling Problem in the Healthcare System. J Healthc Eng. 2022;2022. https://doi.org/10.1155/2022/1938719.

33. Physical properties of almond nut and kernel and modeling the dimensional properties. n.d. https://www.researchgate.net/publication/236856142_Physical_properties_of_almond_nut_and_kernel_and_modeling_the_dimensional_properties. Accessed January 17, 2024.

34. Sharifi, A., Beris, A. T., Javidi, A. S., Nouri, M. S., Lonbar, A. G., & Ahmadi, M. (2024). Application of artificial intelligence in digital twin models for stormwater infrastructure systems in smart cities. Advanced Engineering Informatics, 61, 102485.

35. Lonbar, A. G., Hasanzadeh, H., Asgari, F., Naeini, H. K., Shomali, R., & Asadi, S. (2024). Prioritizing Risk Factors in Media Entrepreneurship on Social Networks: Hybrid Fuzzy Z-Number Approaches for Strategic Budget Allocation and Risk Management in Advertising Construction Campaigns. arXiv preprint arXiv:2409.18976.

36. Reihanifar, M., Takallou, A., Taheri, M., Lonbar, A. G., Ahmadi, M., & Sharifi, A. (2024). Nanotechnology advancements in groundwater remediation: A comprehensive analysis of current research and future prospects. Groundwater for Sustainable Development, 27, 101330.

37. Hussain, A., Reihanifar, M., Niaz, R., Albalawi, O., Maghrebi, M., Ahmed, A. T., & Danandeh Mehr, A. (2024). Characterizing Inter-Seasonal Meteorological Drought Using Random Effect Logistic Regression. Sustainability, 16(19), 8433.

38. Moghim, S., & Takallou, A. (2023). An integrated assessment of extreme hydrometeorological events in Bangladesh. Stochastic Environmental Research and Risk Assessment, 37(7), 2541-2561.

39. Ghoreishi, S. G. A., Moshfeghi, S., Jan, M. T., Conniff, J., Yang, K., Jang, J., ... & Zhai, J. (2023, December). Anomalous behavior detection in trajectory data of older drivers. In 2023 IEEE 20th

International Conference on Smart Communities: Improving Quality of Life using AI, Robotics and IoT (HONET) (pp. 146-151). IEEE.

40. Asadi, S., Gharibzadeh, S., Zangeneh, S., Reihanifar, M., Rahimi, M., & Abdullah, L. (2024). Comparative Analysis of Gradient-Based Optimization Techniques Using Multidimensional Surface 3D Visualizations and Initial Point Sensitivity. arXiv preprint arXiv:2409.04470.

41. Danandeh Mehr, A., Reihanifar, M., Alee, M. M., Vazifehkhah Ghaffari, M. A., Safari, M. J. S., & Mohammadi, B. (2023). VMD-GP: A New Evolutionary Explicit Model for Meteorological Drought Prediction at Ungauged Catchments. Water, 15(15), 2686.